\documentclass[10pt,twocolumn,letterpaper]{article}

\usepackage{cvpr}
\usepackage{lipsum}
\usepackage{times}
\usepackage{epsfig}
\usepackage{graphicx}
\usepackage{amsmath}
\usepackage{amssymb}
\usepackage{booktabs}
\usepackage{color}
\usepackage{subfig}
\usepackage{comment}
\usepackage{enumerate}
\usepackage{arydshln}
\usepackage{rotating}
\usepackage{adjustbox} 
\usepackage[]{algorithm2e}
\usepackage{etoolbox}
\usepackage{multirow}

\makeatletter
\@namedef{ver@everyshi.sty}{}
\makeatother
\usepackage{tcolorbox}

\newcommand{\model}[2]{$\text{#1}_{\text{#2}}$}
\newcommand{\vect}[1]{{\mathbf #1}}
\newcommand{\hstate}[2]{$\vect{h}_{\text{#1}}^{\text{#2}}$}

\usepackage[pagebackref=true,breaklinks=true,letterpaper=true,colorlinks,bookmarks=false]{hyperref}

\cvprfinalcopy % *** Uncomment this line for the final submission

\def\cvprPaperID{1765} % *** Enter the CVPR Paper ID here

\ifcvprfinal\fi
\begin{document}

%%%%%%%%% TITLE
\title{Object Referring in Videos with Language and Human Gaze}

\author{Arun Balajee Vasudevan$^{1}$, Dengxin Dai$^{1}$, Luc Van Gool$^{1,2}$  \\
ETH Zurich$^{1}$ \hspace{1.5cm} KU Leuven $^{2}$  \\
{\tt\small \{arunv,dai,vangool\}@vision.ee.ethz.ch}
%\and
}
\def\AB#1{\color{blue}{#1 }\color{black}} % For Arun's comments and remarks
\def\DD#1{\color{blue}{#1 }\color{black}} % For Dengxin's comments and remarks
\maketitle
%\thispagestyle{empty}

%%%%%%%%% ABSTRACT
\begin{abstract}
We investigate the problem of object referring (OR) \ie to localize a target object in a visual scene coming with a language description. Humans perceive the world more as continued video snippets than as static images, and describe objects not only by their appearance, but also by their spatio-temporal context and motion features. Humans also gaze at the object when they issue a referring expression. Existing works for OR mostly focus on static images only, which fall short in providing many such cues. This paper addresses OR in videos with language and human gaze. To that end, we present a new video dataset for OR, with $30,000$ objects over $5,000$ stereo video sequences annotated for their descriptions and gaze. We further propose a novel network model for OR in videos, by integrating appearance, motion, gaze, and spatio-temporal context into one network. Experimental results show that our method effectively utilizes motion cues, human gaze, and  spatio-temporal context. Our method outperforms previous OR methods. For dataset and code, please refer \url{https://people.ee.ethz.ch/~arunv/ORGaze.html}.

\end{abstract}

%%%%%%%%% BODY TEXT
%\section{Introduction}
%\input{introduction.tex}

\vspace{-2mm}
\section{Introduction}
\label{sec:intro}

In their daily communication, humans refer to objects all the time. The speaker issues a referring expression and the co-observers identify the object referred to. In reality, co-observer also verifies by watching the gaze of the speaker. Upcoming AI machines, such as cognitive robots and autonomous cars, are expected to have the same capacities, in order to interact with their users in a human-like manner.  This paper investigates the task of object referring (OR) in videos with language and human gaze.    

OR has received increasing attention in the last years. Notable examples are interpreting referring expressions \cite{yu2016modeling,mao2016generation}, phrase localization \cite{phloc,wang2016structured}, and grounding of textual phrases \cite{rohrbach2016grounding}. Thanks to these excellent works, OR could be pushed to large-scale datasets~\cite{kazemzadeh2014referitgame, mao2016generation, yu2016modeling} with sophisticated learning approaches~\cite{mao2016generation,hu2016natural,Nagaraja2016, phloc,wang2016structured}. However, previous OR methods are still limited to static images, whereas humans are well aware of the world's dynamic aspects. 
We describe objects not only by their appearance, but also their spatial-temporal contexts and motion features, such as \emph{`the car in front of us turning left'; `the boy running fast under the tree there'}. 
%This can be noted in Fig.~\ref{fig:teaser} also, where \emph{crossing} ascribe to the same  
Static images fall short in providing many of such cues. Thus, there is a strong need to push s-o-a OR to videos. 

\begin{figure}[!tb]
  \centering
  \includegraphics[width=0.47\textwidth]{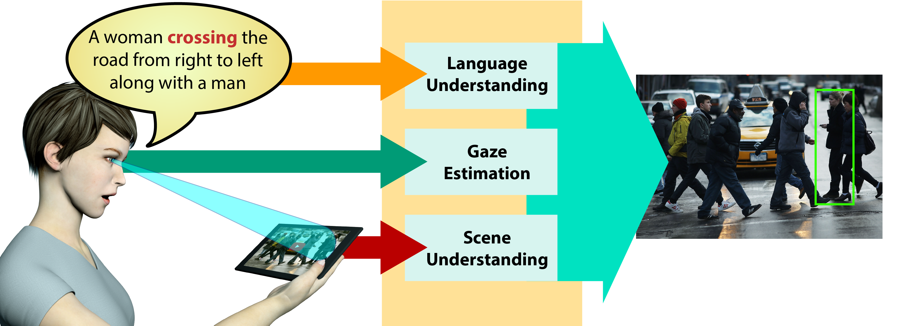}% picture filename
   \caption{A human issuing a referring expression while gazing at the object in the scene. The system combines multiple modalities such as appearance, motion and stereo depth from the video, along with the expression and the gaze to localize the object.}
  \label{fig:teaser}
\end{figure}

Another important cue that co-observers use to identify the objects is Gaze of the speaker. While describing the object, speakers gaze at the object to come out with an unique expression. Gaze is another important cue for object localization from the point of view of co-observer, along with the language expression. For example, suppose a car occupant instructs his/her autonomous car with expression \emph{`Park under the yellow tree on the right'}, it is highly likely that he/she is gazing at that \emph{tree}, or did so in the brief past. This gaze cue can be a promising aid to the car to localize the \emph{tree}. In this work, we also investigate how gaze can be useful in assisting the OR task. As shown in Fig.~\ref{fig:teaser}, we use text language, gaze estimates, visual appearance, motion features and depth features to localize the object being referred.

%This paper investigates OR in videos with language and human gaze. 
As shown several times in computer vision, large-scale datasets can play a crucial role in advancing research, for instance by enabling the deployment of more complex learning approaches and by benchmarking progress. This paper presents a video dataset for OR, the first of its kind, with $30,000$ objects in $5,000$ stereo video sequences. The dataset is annotated with the guidance of  
Gricean Maxims~\cite{Logic:conversation} for cooperative conversations between people. 
That is, the descriptions need to be truthful, informative, relevant, and brief for co-observers to find the target objects easily and unambiguously. Later, human gazes are recorded as videos while they look at the annotated objects.   

\begin{figure*}[t]
\includegraphics[width=1\linewidth,trim={.00\textwidth} {.02\textwidth} {0.0\textwidth} {.01\textwidth},clip]{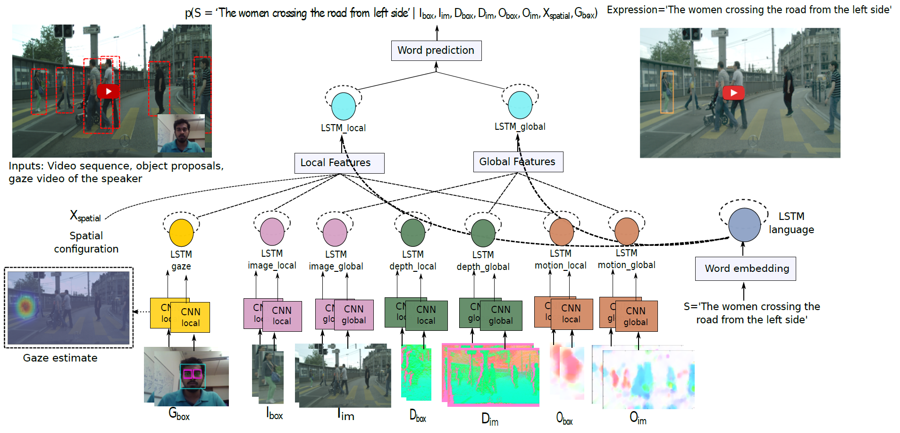}
\caption{The illustrative diagram of our model for object referring in stereo videos with language expression and human gaze. Given a referring expression $S$, our model scores all the $M$ bounding box candidates by jointly considering local appearance ($I_{\text{box}}^t$), local motion ($O_{\text{box}}^t$), local depth ($D_{\text{box}}^t$), local human gaze ($G_{\text{box}}^t$), spatial configuration $X_{\text{spatial}}$, and the global temporal-spatial contextual information ($I^t$, $D^t$ and $O^t$).}
\label{fig:pipeline}
\end{figure*}

We further propose a novel Temporal-Spatial Context Recurrent ConvNet model, by integrating appearance, motion, gaze, and spatial-temporal context into one network. See Fig.~\ref{fig:pipeline} for a diagram of our model. The model learns the interactions between language expressions and object characteristics in the `real' 3D world, providing human users the freedom to interact by speaking and gazing.  Experimental results show that our method effectively uses motion cues, temporal-spatial context information, and human gazes.

Our main contributions are: 1) presenting a new video dataset for object referring, featuring bounding-boxes, language descriptions and human gazes; 2) developing a novel OR approach to detect objects in videos by learning from appearance, motion, gaze, and temporal-spatial context. 

%------------------------------------------------------------------------
\vspace{-1mm}
\section{Related Work}
\label{sec:related} 

Our work is relevant to the joint understanding of language and visual data. It is especially relevant to referring expression generation and language-based object detection.

The connection between language and visual data has been extensively studied in the last three years. The main topics include image captioning~\cite{deep:semantic:alignment:cvpr15, show:attend:tell, caption:back}, visual question answering (VQA)~\cite{exploraing:models:data:vqa, VQA, andreas2016learning} and referring expressions~\cite{hendricks2017localizing,hu2016natural}. Although the goals are different, these tasks share many fundamental techniques. Two of the workhorses are  Multimodal Embedding~\cite{frome2013devise, deep:semantic:alignment:cvpr15, multimodal:pooling} and Conditional LSTM~\cite{densecap:cvpr16, rohrbach2016grounding, yu2016modeling, mao2016generation}.  Multimodal Embedding projects textual data and visual data both to a common space, in which similarity scores or ranking functions are learned.  
%A common representation space allows for discriminative learning of the interactions between the two modalities. 
Multimodal Embedding was initially explored for the task of image captioning~\cite{frome2013devise, long-term:recurrent, deep:semantic:alignment:cvpr15} and later reinforced in VQA~\cite{exploraing:models:data:vqa,ask:your:neurons,multimodal:pooling}. 
It is common practice to represent visual data with CNNs pre-trained  for image recognition and to represent textual data with word embeddings pre-trained on large text corpora~\cite{glove}. A Conditional LSTM is a generative model conditioned on visual input, and it is usually trained to maximize the generative probability of language descriptions~\cite{hu2016natural,mao2016generation} or answers to questions \cite{VQA,exploraing:models:data:vqa}. %In principle, a Conditional LSTM is more flexible and expressive, and thus more suitable for tackling open-ended language-vision problems. 
Our model conditions LSTMs not only on images but also on motion, depth and gaze.

\vspace{-2mm}
\bigskip
\noindent
\textbf{Language-based Object Referring.}
Language-based object referring (OR) has been tackled under different names.  
Notable ones are referring expressions \cite{yu2016modeling,mao2016generation}, phrase localization \cite{phloc,wang2016structured}, grounding of textual phrases \cite{rohrbach2016grounding,yehNIPS17,chen2017query}, language-based object retrieval~\cite{hu2016natural} and segmentation~\cite{hu2016segmentation}. 
Recent research foci of language based OR can be put into 2 groups: 1) learning embedding functions~\cite{multimodal:pooling,deep:semantic:alignment:cvpr15, wang2016learning} for effective interaction between vision and language; 2) modeling contextual information to better understand a speaker's intent, be it global context~\cite{mao2016generation,hu2016natural}, or local among `similar' objects~\cite{Nagaraja2016,yu2016modeling,mao2016generation}. Our work extends \cite{hu2016natural} from static images to stereo videos to exploit richer, more realistic temporal-spatial contextual information along with gaze cues for the task of OR.

\vspace{-2mm}
\bigskip
\noindent
\textbf{Object Referring Datasets.} 
This section discusses relevant OR datasets:    
Google Refexp~\cite{mao2016generation}, UNC Refexp~\cite{yu2016modeling}, ReferIt~\cite{kazemzadeh2014referitgame}. 
The Google Refexp dataset, which was collected by Mao~\etal~\cite{mao2016generation}, contains $104,560$ referring expressions annotated for $54,822$ objects from $26,711$ images from the MSCOCO dataset~\cite{lin2014microsoft}. UNC Refexp was collected in the same spirit as GoogleRef, but applying the ReferIt game~\cite{kazemzadeh2014referitgame} on MSCOCO. While these datasets are large-scale and of high quality, they contain only static images. This excludes useful information about the visual scenes such as motion cues and 3D spatial configuration, and also limits the descriptions to mere appearances and 2D spatial information. We build on the success of these datasets and present a new object referring dataset for stereo videos. Annotators were encouraged to use descriptions about 3D configuration and motion cues when necessary. 

\vspace{-2mm}
\bigskip
\noindent
\textbf{Gaze Estimation.} Gaze or eye tracking has been used in computer vision tasks like object detection~\cite{karthikeyan2013and,yun2013studying} and tracking~\cite{Vadivel_2015_CVPR}, image captioning~\cite{sugano2016seeing}, image/video annotation ~\cite{Vadivel_2015_CVPR, papadopoulos2014training, soliman2016towards} and others. We focus on object referring. There are some works~\cite{kreysa2011peripheral, staudte2014influence} which support that speaker’s gaze reliably precedes his reference to an object and this speaker's referential gaze helps in listeners' comprehension. Common sense also tells us that we have to gaze at the object before we refer them. Listeners use this gaze information for a better object localization ~\cite{staudte2014influence, nunnemannreferential, knoeferle2012can}. Misu~\etal~\cite{misu2013situated} uses gaze and speech recorded inside a car to locate real world landmark points. Vadivel~\etal~\cite{Vadivel_2015_CVPR} uses eye tracking over videos to extract salient objects as tracklets. The task of \cite{papadopoulos2014training} matches closely with us to detect objects using gaze. Nonetheless, they use gaze to fasten the annotation for object detection. \cite{nips15_recasens} proposes to follow the gaze of people in the scene to localize the place where they look. Krafka~\etal~\cite{krafka2016eye} build an mobile application to enable large scale eye tracking annotation by crowdsourcing. Inspired from \cite{krafka2016eye}, we create a web interface to record gaze via Amazon Mechanical Turk (AMT) for object localization in videos. 

\vspace{-1.5mm}
\section{Approach} 
\label{sec:approach}
Object referring (OR) is widely used in our daily communication. Here, we follow the literature~\cite{hu2016natural,mao2016generation} to formulate the problem as an object detection task. Given a video sequence of visual scene $\mathbf{I}=(I^1, I^2, ...,I^t)$ and a video sequence of speaker's gaze $\mathbf{G}=(G^1, G^2, ...,G^t)$, where $t$ is the \emph{current} frame at which the referring expression $S$ is issued, our goal is to identify the referred object $\hat{b}^t$ out of all object proposals $\{b_m^t\}_{m=1}^M$  at frame $t$. $M$ is the total number of object proposals considered. Note that we assume that $t$ is known a priori to simplify the task. In real application, the exact $t$ needs to be inferred from speaker's speech and the visual scene. The performance of our method is also evaluated at frame $t$.

\subsection{Network Architecture} 
\label{sec:model}

Following \cite{hu2016natural} and the work on image captioning~\cite{densecap:cvpr16}, we choose to maximize the generative probability of the expression for the target object.   Our model is based on the Spatial Context Recurrent ConvNet  model developed in ~\cite{hu2016natural} for OR in static images. The model in~\cite{hu2016natural}  unifies three LSTMs~\cite{lstm} to integrate information from language expressions, global visual context and local object content. 
It has gained success in OR for static images. This work extends it so that information from stereo videos and human gazes can be incorporated, resulting in our model architecture as shown in Fig.~\ref{fig:pipeline}.

Let us denote the seven \textbf{visual} LSTM models by \model{LSTM}{gaze}, \model{LSTM}{image\_local}, \model{LSTM}{image\_global}, \model{LSTM}{depth\_local}, \model{LSTM}{depth\_global}, \model{LSTM}{motion\_local} and \model{LSTM}{motion\_global}, and their hidden states by \hstate{}{gaze}, \hstate{local}{image}, \hstate{global}{image}, ..., \hstate{global}{motion}, respectively and denote the \textbf{language} LSTM model by \model{LSTM}{language} with hidden state \hstate{}{language}. We concatenate the local and global features separately as shown in Fig.~\ref{fig:pipeline}. Successively, we have \textbf{visual-language} LSTM models namely \model{LSTM}{local} and \model{LSTM}{global} which take concatenated local and global features respectively along with \hstate{}{language} as inputs. Let us denote their hidden states as \hstate{local}{} and \hstate{global}{} respectively.
A word prediction layer is used on top of these two visual-language LSTMs to predict the words in the expression $S$. Practically, our model is trained to predict the conditional probability of the next word $w_{n+1}$ in $S$,  given the local content of the objects: $G_{\text{box}}^t$, $I_{\text{box}}^t$, $D_{\text{box}}^t$ and $O_{\text{box}}^t$, the corresponding spatio-temporal contexts: $I^t$, $D^t$ and $O^t$ as detailed in Sec.~\ref{sec:featureencod}, and all the $n$ previous words. The problem can be formulated as: 
\begin{multline}
p(w_{n+1}|w_n, ..., w_{1}, I^t, I_{\text{box}}^t, D^t, D_{\text{box}}^t, O^t, O_{\text{box}}^t, G_{\text{box}}^t) \\
= \text{SoftMax}(W_{\text{local}}\vect{h}_{\text{local}}(n) + W_{\text{global}}\vect{h}_{\text{global}}(n) + \vect{r})
\end{multline}
where $W_{\text{local}}$ and $W_{\text{global}}$ are the weight matrices for word prediction from \model{LSTM}{local} and \model{LSTM}{global}, and $\vect{r}$ is a bias vector.  

At training time, the method maximizes the probability of generating all the annotated expressions over the whole dataset. Following \cite{hu2016natural}, all the seven LSTM models have $1000$ hidden states. 
At test time, given a video sequence $\mathbf{I}$, a gaze sequence $\mathbf{G}$ and $M$ candidate bounding boxes $\{b_m^t\}_{m=1}^M$ at frame $t$ considered by the method proposed in Sec.~\ref{sec:objprop}, our model computes the OR score for $b_m$ by computing the generative probability of $S$ on $b_m^t$(box):
\begingroup\makeatletter\def\f@size{8}\check@mathfonts
\begin{multline}
s_i=p(S|I^t, I_{\text{box}}^t, D^t, D_{\text{box}}^t, O^t, O_{\text{box}}^t, G_{\text{box}}^t) \\
=\prod_{w_n \in S}{p(w_n|w_{n-1}, ..., w_1, I^t, I_{\text{box}}^t, D^t, D_{\text{box}}^t, O^t, O_{\text{box}}^t, G_{\text{box}}^t ).} 
\end{multline}
\endgroup
The candidate with the highest score is taken as the predicted target object. Below, we describe our object proposal and feature encoding. 

\subsection{Object Proposals}
\label{sec:objprop} 

In the spirit of object detection, we adopt the strategy of proposing candidates efficiently and then verifying the candidates with a more complex model for the OR task. This strategy has been used widely in the literature. For instance, \cite{hu2016natural} uses EdgeBox~\cite{ZitnickECCV14edgeBoxes} for the object proposals;  
\cite{mao2016generation} and \cite{deep:semantic:alignment:cvpr15} use the faster RCNN (FRCNN) object detector \cite{renNIPS15fasterrcnn}, Mask-RCNN~\cite{he2017mask} and Language based Object Proposals (LOP)~\cite{vasudevan2017chi} and others to propose the candidates. 
%It has been shown that using proposals strikes a good trade-off between accuracy and speed. 
\cite{vasudevan2017chi} shows that LOP performs significantly better than other techniques when we propose expression-aware object candidates. For the same reason, we use LOP~\cite{vasudevan2017chi} for the object proposals.

\subsection{Feature Encoding}
\label{sec:featureencod} 
In order to better use the spatio-temporal information provided by a stereo video, we augment $I^t$ with the corresponding depth map $D^t$ and optical flow map $O^t$. In addition to these global contexts, for a bounding box $b^t$, its local features are used as well: $I_{box}^t$ for its appearance, $D_{box}^t$ for its depth characteristics, $O_{box}^t$ for its motion cues and $G_{box}^t$ for the gaze. CNNs are used to encode the local and global information from the three information sources. $I_{box}^t$, $D_{box}^t$ and $O_{box}^t$ can be computed on frame $t$ alone or together with multiple previous frames for long-range temporal interaction. The same is applicable for $I^t$, $D^t$ and $O^t$ also. The detailed evaluation can be found in Sec. \ref{sec:experiment}.  

\noindent \textbf{Appearance.}
We use the \emph{fc7} feature of VGG-16 net~\cite{vgg16} and ResNet~\cite{he2016deep} pre-trained on ILSVRC-2015~\cite{imagenet:2015} to represent $I_{box}^t$ and $I^t$, which are passed through \model{LSTM}{image\_local} and \model{LSTM}{image\_global} respectively to yield features $\vect{f}_{local}^{image}$ and $\vect{f}_{global}^{image}$, respectively.

\noindent \textbf{Depth.}
For depth, we convert depth maps to HHA images~\cite{rgbd:net:eccv14} and extract the CNN features with the RGB-D network of ~\cite{rgbd:net:eccv14} before passing to \model{LSTM}{depth\_local} and \model{LSTM}{depth\_global}. This leads to  depth features $\vect{f}_{local}^{depth}$ and $\vect{f}_{global}^{depth}$ for $D_{box}^t$ and $D^t$, respectively. 

\noindent \textbf{Optical Flow.}
Similarly, we employ the pre-trained two-stream network~\cite{two:stream:nips14} trained for video action recognition to extract convolutional flow features. Again, the \emph{fc7} features are used leading to $4096$-dimensional features which are given to \model{LSTM}{motion\_local} and \model{LSTM}{motion\_global} to get the motion features $\vect{f}_{local}^{motion}$ and $\vect{f}_{global}^{motion}$ for $O_{box}^t$ and $O^t$. 

\noindent \textbf{Language.} 
The words in the expression $S$  are represented as one-hot vectors and embedded by word2vec~\cite{glove} first and later, the expression $S$ is embedded by an LSTM model~\cite{lstm} \model{LSTM}{language}, leading to a language feature vector $\vect{h}^{\text{language}}$ for $S$.  

\noindent \textbf{Human Gaze.}
We synchronize the video of human gaze and the Cityscapes's video, which was displayed on a laptop for gaze recording. On the extracted frames, we perform face detection to crop out the face image and then conduct facial landmark point detection using the work of \cite{kowalski2017deep}. Successively, we detect the left eye and the right eye, and extract them as well. An example is shown in Fig.~\ref{fig:pipeline}. Then, we use GazeCapture model~\cite{krafka2016eye} which takes input of left and right eye images along with the face image and outputs the gaze estimates relative to the camera location on the device. We convert the estimate from camera coordinate system to image coordinate system by applying a linear mapping function. The mapping function is device-specific and defined by the relative position of the camera and the screen of the laptop. More details of the mapping function can be found in the supplementary material. 

Finally, we plot a 2D Gaussian map around the estimated gaze coordinates on the image to accommodate the errors in gaze estimation by the GazeCapture model~\cite{krafka2016eye} as shown in Fig.~\ref{fig:gaze_error}. Later, we compute gaze feature for an object in a frame by region pooling (averaging) over the bounding box region inside the Gaussian map. We concatenate these features over all the frames of the video to yield a resultant gaze feature $\vect{f}^{gaze}$ for an object~\ie $G_{box}^t$.

\noindent \textbf{Feature Concatenation.}
Similar to \cite{hu2016natural}, we concatenate the language feature with each of the two concatenated local and global features to obtain two meta-features: $\{[\vect{h}^{\text{language}}, \vect{f}_{local}^{image}, \vect{f}_{local}^{depth}, \vect{f}_{local}^{motion}, \vect{f}^{gaze},\vect{f}_{spatial}]$, $[\vect{h}^{\text{language}}, \vect{f}_{global}^{image}, \vect{f}_{global}^{motion}, \vect{f}_{global}^{motion}]\}$. The meta-features are then given as the input to two LSTM models \model{LSTM}{local} and \model{LSTM}{global} respectively, to learn the interaction between language and all the `visual' domains, \ie appearance, motion, depth, gaze and their global contexts.  
$\vect{f}_{spatial}$ denotes the spatial configuration of the bounding box  with respect to the 2D video frame. Following~\cite{mao2016generation,hu2016natural}, this $8-$dimensional feature is used:  
\begin{equation}
\vect{f}_{\text{spatial}}=[x_{\text{min}}, y_{\text{min}}, x_{\text{max}}, y_{\text{max}}, x_{\text{center}}, y_{\text{center}}, w_{\text{box}}, h_{\text{box}}]
\end{equation}
where $w_{\text{box}}$ and $h_{\text{box}}$ are the width and height of the box. 
%$\vect{f}_{\text{spatial}}$ is added to the augmented local image features.%, leading to a new feature $[\vect{h}_{\text{language}}, \vect{f}_{image}^{local}, \vect{f}_{\text{spacial}}]$ for the model \model{LSTM}{image\_local}. 
See Fig.~\ref{fig:pipeline} for all the interactions. Our method can be trained in a full end-to-end fashion if enough data is provided.  

\begin{figure}[t]
\centering
\begin{tabular}{lll}
\includegraphics[width=0.5\linewidth]{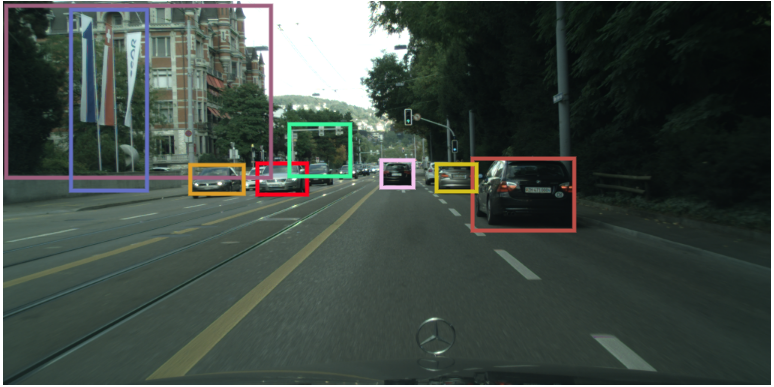}
\includegraphics[width=0.5\linewidth]{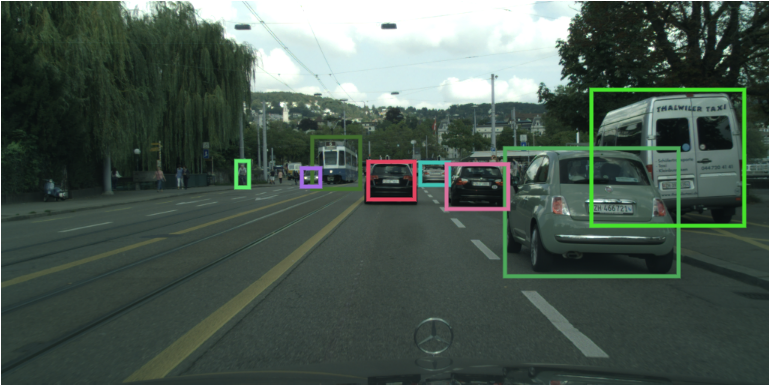}
\noindent 
\noindent
\\
\noindent 
\includegraphics[width=0.5\linewidth,height=0.50\linewidth]{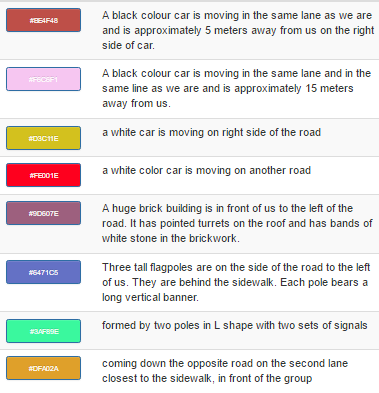}

\includegraphics[width=0.5\linewidth,height=0.50\linewidth]{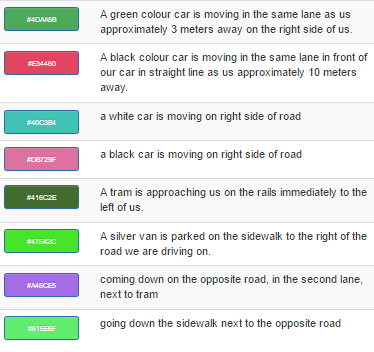}
%\noindent 
%\includegraphics[width=0.23\linewidth,height=0.20\linewidth]{figures/desc3.PNG}
%\noindent 
%\includegraphics[width=0.23\linewidth,height=0.20\linewidth]{figures/desc4.PNG}

\end{tabular}
\caption{Top: sample images from the Cityscape dataset with objects marked in differently colored bounding boxes. Bottom: corresponding referring expression annotations.}
\label{fig:recallvsiou}
\end{figure}

\section{Dataset Annotation}
\label{sec:dataset}

As discussed in Sec.~\ref{sec:intro} and Sec.~\ref{sec:related}, previous datasets do not cater to the learning and evaluation of temporal, spatial context, and gaze information.  Thus, we collected a new dataset.  A video OR dataset should contain  diverse visual scenes and their objects should be annotated at the frame (time) when the expression is issued. We acknowledge that all modalities should be recorded/annotated at the same time, ideally in the real human-to-robot communication scenarios. That, however, renders data collection very labor-intensive and infeasible to crowd source.  

In this work, we choose to use the existing stereo videos from Cityscapes dataset~\cite{cityscape}, and annotate language expressions, object bounding boxes, and gaze recordings via crowd sourcing. Cityscapes consists of $5,000$ high-quality video sequences in total, captured with a car mounted stereo camera system in $50$ different European cities.  The videos contain diverse sets of traffic scenes such as \emph{car approaching a signal stop}, \emph{pedestrians crossing the road}, \emph{trams running through the street}, and \emph{bicycles are overtaken}, and \emph{kids crossing  road lanes}, \etc. See Fig.~\ref{fig:recallvsiou} for some examples.

\bigskip
\noindent
\textbf{Crowdsourcing}. We crowdsourced the annotation task of OR in videos via AMT. Each Human Intelligence Task (HIT) contains one video.
The videos in the Cityscapes dataset are all 2 seconds long, comprising 30 frames. An AMT worker was asked to annotate bounding boxes for objects on the last frame of the video (\ie the 30th frame). The 30th frame is chosen mainly to  make sure that annotated objects come with sufficient temporal context. In the annotation, workers are `forced' to watch the video at least once in order to annotate an object. Replaying the video is highly encouraged if something is unclear.

\bigskip
\noindent
\textbf{Quality Control}. To generate high quality annotations, we ran the first round of HIT as a qualification task. We qualified 20 workers based on their annotation of bounding boxes and natural language descriptions who further annotated the entire dataset. %The annotations are also manually checked to spot erroneous annotations. 
Following the work of Li~\etal \cite{li2016tgif}, we employed various quality control mechanisms for the syntactic validation to ensure the high quality of sentences. Some of the used validation checks are: number of words in the description must be at least 5, words must contain only ASCII characters, copy/paste operation is not allowed in the field where workers typed the descriptions and finally, we check for grammatical and spelling errors using the HTML5 spellcheck attribute. We payed $0.075$ US dollar for each annotation of one bounding box, the name of the object class, and a referring expression. In total, we have collected $30,000$ annotated objects in $5,000$ videos.  %The annotation interface will be provided in the supplementary material. 

\bigskip
\noindent
\textbf{Dataset Statistics}. The average length of referring expressions of the objects is $15.59$ words compared to $8.43$ in Google Refexp and $3.61$ in the UNC Refexp dataset, which are popular referring expression datasets. 
%Our expressions are richer mainly because the scenes are more complex and the speakers (Workers) need to distinguish the referred object from other traffic agents. 
There are 20 classes of objects in Cityscape. The average number of referring expressions on objects annotated per image is $4.2$ compared to $3.91$ in Google Refexp. The distribution of annotations is 53.99\% referring expressions for 'car', 22.97\% for 'person', 4.9\% for 'vegetation', 3.9\% for 'bicycle', 3.46\% for 'building', 2.95\% for 'rider' and the rest for the remaining categories. %There was a large difference in the number of annotations for each object.

\bigskip
\noindent
\textbf{Gaze Recording}. As a separate annotation task, we record human gaze for the objects which have been annotated already with referring expressions and bounding boxes. This is inspired from Krafke~\etal \cite{krafka2016eye} where eye tracking dataset is created via crowdsourcing by asking workers to gaze at particular points on the device screen with their face being recorded. Here, we collect the gaze recording on objects annotated in Cityscapes dataset as mentioned beforehand in this section. We create a web interface where we show the videos and asked the turk workers to gaze at the object one after the other. We record the faces of workers using the frontal camera of their laptops while they gaze at the shown objects.

In our annotation interface, we instruct the workers to adjust the window size such that the canvas where videos are displayed, occupies a major amount of screen space for a higher resolution of gaze estimation. 
With the start of the annotation, workers are asked to watch the complete video at first to put them into context.
Once this is done, we show the objects (in bounding boxes) with annotated bounding boxes on the canvas. The workers are asked to click inside the box which activates the running of the video. 
We direct the workers to gaze at the same clicked object during the stage of video streaming while we keep recording the gaze of the worker throughout this period. Successively, we show the next objects and record the gazes correspondingly. 
At the end of the annotation of each video, we collect videos for the gaze recording of every annotated object.

We ensured the quality of recording by allowing only qualified workers to participate in this task. We perform the qualification same as in the earlier task except the criteria being checked is \textit{gazing} here. Workers perform the task under different lighting conditions and at times, their visibility of their face goes down with its consequence being that the face detection fails in those cases. Hence, we re-recorded the gazes for all the videos where face detection on the workers failed. Finally, we recorded gaze for all the annotated objects of the Cityscapes.

\section{Experiments}
\label{sec:experiment}

Given a video sequence of visual scene, gaze recording sequence of the speaker and a referring expression, our task is to yield bounding box location of the object. 
Our model scores and ranks the object proposals (which we generate using LOP~\cite{vasudevan2017chi}) based on the textual description, the spatial and temporal contextual information from stereo videos and gaze recording of the speaker. We first evaluate the performance of multiple modalities namely, RGB image, depth information and object motion. It is aimed to show the usefulness of depth and motion provided by stereo videos for the task of Object Referring (OR). Later, we show how gaze aids our model to improve the OR accuracy further.

\subsection{Implementation Details}
Our model is designed to incorporate gaze of the speaker, temporal and depth cues of the objects as well as the contextual information. The main advantage of Cityscapes referring expression annotations over other referring expressions datasets like GoogleRef, UNC Refexp and ReferIt is that the Cityscapes consists of short video snippets and the corresponding depth maps, suited for our task. For RGB images, we extract features from VGG16~\cite{vgg16} and ResNet~\cite{he2016deep} as mentioned in Sec.~\ref{sec:approach}. For depth, we generate HHA images from disparity maps following the work of Gupta~\etal~\cite{rgbd:net:eccv14}. Furthermore, we extract HHA features using the RCNN network used by \cite{rgbd:net:eccv14}. For motion, we compute optical flow for all the frames using Fast Optical Flow by Kroeger~\etal~\cite{kroeger2016fast}. We extract optical flow features using the flow network of the two stream Convolutional network implemented by Simonyan~\etal~\cite{two:stream:nips14} for action recognition in videos. To compute object level features in all frames of videos, we compute tracks of the objects using the annotated bounding box on the last frame of the videos(30th frame in Cityscapes). We compute tracks for each object using Correlation filter based tracking~\cite{valmadre2017end}. 

\begin{figure}[t]
\centering
\begin{tabular}{ll}
\includegraphics[width=0.5\linewidth,trim={.22\textwidth} {.05\textwidth} {0.12\textwidth} {.05\textwidth},clip]{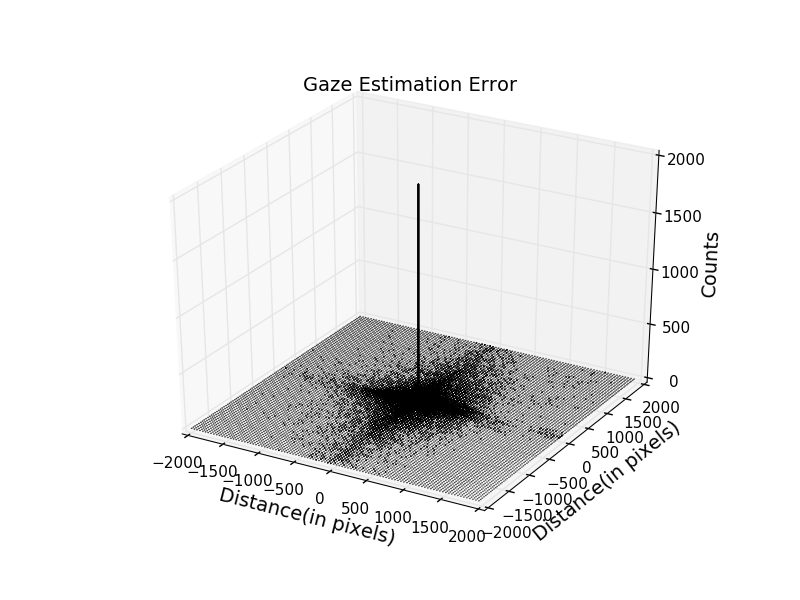}
\includegraphics[width=0.5\linewidth,trim={.05\textwidth} {.00\textwidth} {0.132\textwidth} {.05\textwidth},clip]{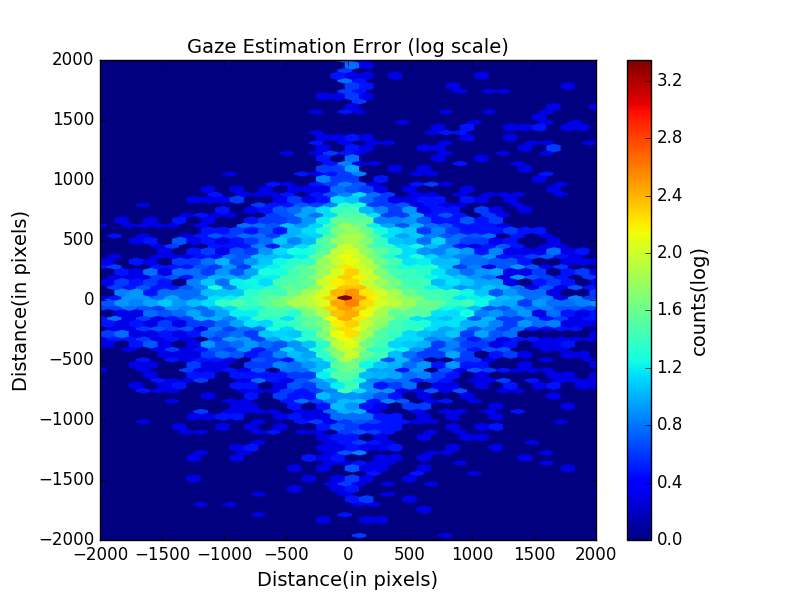}
%\noindent 
%\noindent
\end{tabular}
\vspace{-1mm}
\caption{Gaze Estimation error distribution. We compute the distance between the gaze estimation coordinates with the groundtruth bounding box along X and Y axis(Centre denotes zero error). Left side figure represents the error in real valued scale and right side in log scale. We choose 2000 pixel distance to match with Cityscapes image dimensions.}
\vspace{-3mm}
\label{fig:gaze_error}
\end{figure}

As to Gaze, we sample frames from the gaze video at a frame rate greater than that of the Cityscapes video to ensure one-to-one correspondence between the sequences. Then, we extract the face from each frame with~\cite{viola2001rapid}. For these face images, we use Deep Alignment Network~\cite{kowalski2017deep} to extract facial landmark points. Using the left and right eye landmark points, we extract the left and right eye image which we later give to GazeCapture model~\cite{krafka2016eye} along with face information. GazeCapture model outputs the gaze prediction in camera coordinates. We convert these camera coordinates to image coordinates using the linear mapping function as mentioned in Sec.~\ref{sec:approach}. 
Then, we plot 2D Gaussian plot around the gaze estimate in image coordinates with $\sigma=100$ pixels which is 10\% of image dimension. This helps in accommodating the prediction error of gaze coordinates from the GazeCapture model as mentioned in ~\cite{krafka2016eye}. See Fig.~\ref{fig:gaze_error} for gaze prediction error distribution. Successively, we perform region pooling over the bounding box location of Gaussian map to obtain the object feature. We concatenate the features from each frame to get the gaze feature for the entire recording. Finally, we train our model by using the extracted features from RGB, HHA, Optical Flow images and gaze features as shown in Fig.~\ref{fig:pipeline}.

%-----------------------------------------------------------------------
\begin{table}[t]
\centering
\resizebox{0.9\columnwidth}{!}{
\begin{tabular}{@{}lllllllllllll@{}}
\toprule
 %&            & \multicolumn{2}{c}{HIT @ 1} &   &         \\ \cmidrule{3-4}
 &Methods &\multicolumn{1}{c}{Edgebox}&\multicolumn{1}{c}{FRCNN} &\multicolumn{1}{c}{LOP }\\
% \cmidrule(lr){1-2} \cmidrule(lr){3-3} \cmidrule(lr){4-6}  \cmidrule(l){7-9}
% &Candidate Proposals &\multicolumn{1}{c}{100} &\multicolumn{1}{c}{10} &\multicolumn{1}{c}{30}&\multicolumn{1}{c}{100 }& \multicolumn{1}{c}{10 }  & \multicolumn{1}{c}{30 } & \multicolumn{1}{c}{100 }      \\ 
% \cmidrule(lr){1-2} \cmidrule(lr){3-3} \cmidrule(l){4-5}\cmidrule(l){6-7} \cmidrule(l){8-10}\cmidrule(l){11-13}
%&Method         &Acc@1 &  Acc@1  &  Acc@5 &Acc@1  &  Acc@10 & Acc@1 &  Acc@10  &  Acc@30  & Acc@1 &  Acc@10  &  Acc@100    \\ 
\midrule
\parbox[t]{2mm}{\multirow{5}{*}{\rotatebox[origin=c]{90}{VGG}}}
%\multicolumn{10}{l}{\textsc {GoogleRef}}   \\
%& Random &0.98& 3.75& -  \\
%& NLOR \cite{hu2016natural} & 9.02& -&25.04& -&36.1& -& -&? & 15.01 & 50.61& 73.78 \\
& SimModel \cite{phloc}&4.5& 18.431& 35.556 \\
& MNLM \cite{kiros2014unifying}&-& 23.954 & 32.418 \\
& VSEM \cite{liu2015multi}&-& 24.833 & 32.961 \\
& MCB \cite{multimodal:pooling}&-& 26.445& 33.366 \\
& NLOR \cite{hu2016natural}(Ours(I)) &4.1& 27.150& 36.895 \\

\midrule 
\parbox[t]{2mm}{\multirow{5}{*}{\rotatebox[origin=c]{90}{ResNet}}}
& NLOR-ResNet (Ours(I)) && 29.333& 38.645 \\
%& Mao ~\etal \cite{mao2016generation}  & 17.5 & - &- \\
%& Ours (I,D) && 38.562& 40.948  \\
& Ours (I,D) &-& 38.833& 41.388  \\
%& Ours (I,O) && 38.856& 42.255  \\
& Ours (I,O) &-& 39.166& 42.500 \\
%& Ours (I,D,O) & &39.509& 43.137  \\
& Ours (I,D,O) &- &41.205& 43.750  \\
%& Ours (I,D,O,G) & & \textbf{45.784}& \textbf{45.065}  \\
& Ours (I,D,O,G) &- & \textbf{47.256}& \textbf{47.012}  \\

%& Mao ~\etal \cite{mao2016generation}  & 17.5 & - &- \\
%& Ours (I,D) & -&38.562& 40.948  \\
%& Ours (I,O) & -&38.856& 42.255  \\
%& Ours (I,D,O) & -& 39.509& 43.137  \\
%& Ours (I,D,O,G) & -& \textbf{45.784}& \textbf{45.065}  \\

\bottomrule
\end{tabular}
}
\caption{Numbers denote Acc@1. The \# of candidate proposals $M$ is $30$. All evaluations are on Cityscapes. Abbreviations: I:RGB, D:Depth map, O:Optical Flow, G:Gaze. Since Edgebox performs poorly in baseline:Ours(I), we avoid further experiments.}
\label{tab:eop-expt}
\end{table}

\begin{table}[t]
\centering
\resizebox{0.7\columnwidth}{!}{
\begin{tabular}{@{}llllll@{}}
\toprule
& &\multicolumn{3}{c}{Track length(in frames)}     \\
 \cmidrule(lr){3-5}
&Methods & 1 &  2  &  8     \\ \midrule
& Ours (I) &38.645& -& -  \\
& Ours (I,O)  &42.500& 42.418& 42.320 \\
& Ours (I,D,O)  &\textbf{43.750}&\textbf{42.875}& \textbf{42.875} \\
\bottomrule
\end{tabular}
}
\caption{Comparison of methods when longer term motion is considered. Numbers denote Acc@1. Track length represents the number of past frames used for flow information. The different methods are evaluated on Cityscape.}
\label{tab:cityscape-track}
\end{table}

\begin{table}[t]
\resizebox{1.0\columnwidth}{!}{
\centering
\begin{tabular}{@{}llllll@{}}
\toprule
%& &\multicolumn{3}{c}{Track length(in frames)}     \\
% \cmidrule(lr){3-5}
&Methods & Ours (I) & Ours (I,O) & Ours (I,D,O)      \\ \midrule
& w/o Gaze & 38.645&42.500 & 43.750  \\
%& w/ Gaze +AvgPool & 42.026& 41.895& 42.516  \\
& w/ Gaze +AvgPool & 41.242& 41.895& 43.791  \\
& w/ Gaze & 41.535& 43.888 & 45.816  \\
& w/ Gaze +MaxPool & \textbf{42.418}& \textbf{44.248} & \textbf{47.012}  \\
\bottomrule
\end{tabular}
}
\caption{Comparison of approaches w/ and w/o Gaze. Numbers denote Acc@1. \# of candidate proposals $M$ is 30. The different methods are evaluated on Cityscape. 1st row has overlap with Tab.~\ref{tab:eop-expt}}
\label{tab:gaze-comparison}
\end{table}

\subsection{Evaluation}
Out of the $30,000$ annotated objects in our Cityscape dataset~\cite{cityscape}, we use 80\% of videos for the training and 20\% for the evaluation of our model on the task of OR in videos.

\bigskip
\noindent
\textbf{Evaluation Metric}
We evaluate the performance of our language based OR based on Accuracy@1 ($Acc@K$), following ~\cite{hu2016natural,mao2016generation}. 
%$Acc@K$ refers to the percentage of true detections among $K$ top scored object candidates. 
$Acc@1$ refers to the percentage of top scoring candidates being a true detection. A candidate is regarded as a true detection if the Intersection over Union (IoU) computed between the predicted bounding box and ground truth box is more than $0.5$. 
%If the IoU is less than $0.5$, the detection is considered a false positive. 
In all our tables, we compute mean of the $Acc@1$ metric over all the videos in the evaluation set.

%---------------------------------------------------
\begin{figure}[t!]
\centering
{\tiny \textbf{NLOR}} \hspace{60pt} {\tiny\textbf{Ours(I,D,O)}} \hspace{30pt} {\tiny \textbf{Ours(I,D,O,G)}}
\begin{tabular}{lll}
\multicolumn{3}{c}{\tiny a woman in front with a white cap is walking towards us on right side of the road along with others}\\
\noindent
\adjustbox{trim={.1\width} {.22\height} {0.1\width} {.22\height},clip}
    {\includegraphics[width=0.4\linewidth]{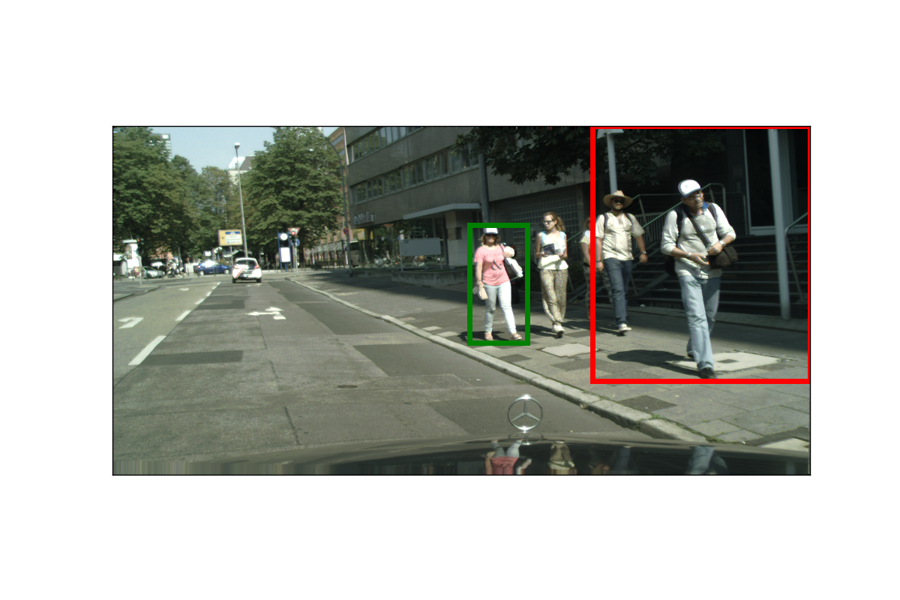}}
\noindent 
\adjustbox{trim={.1\width} {.22\height} {0.1\width} {.22\height},clip}
    {\includegraphics[width=0.4\linewidth]{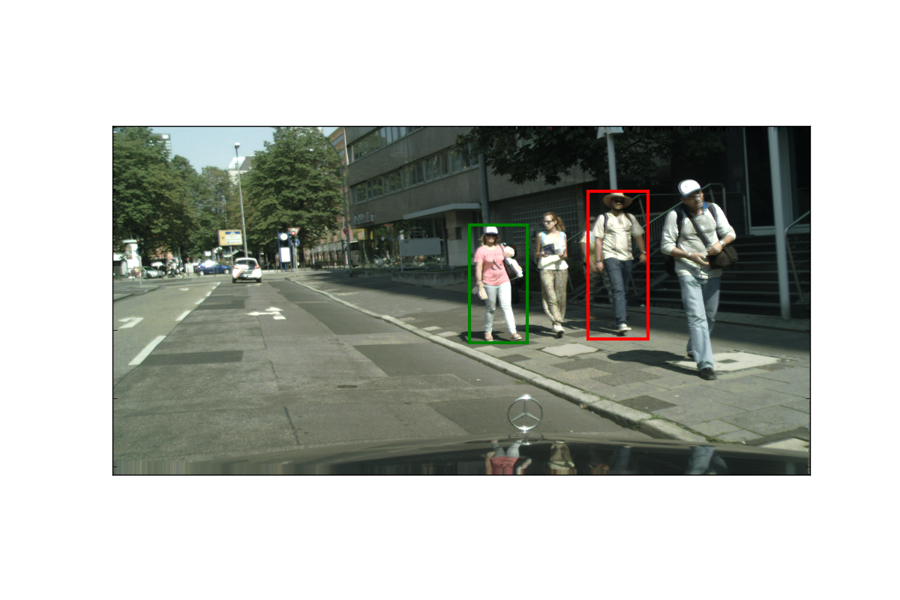}}
    \adjustbox{trim={.1\width} {.22\height} {0.1\width} {.22\height},clip}
    {\includegraphics[width=0.4\linewidth]{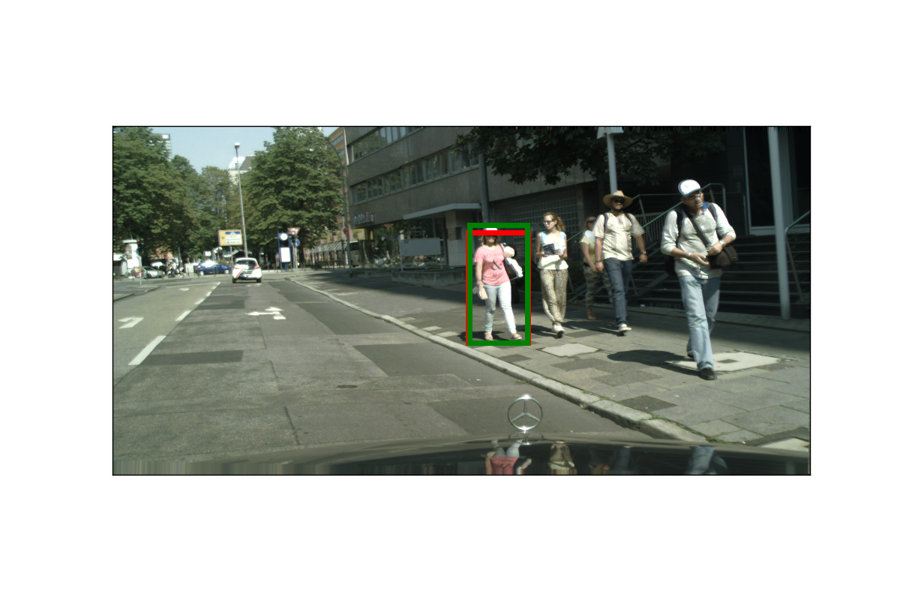}} \vspace{-1mm}\\
    \multicolumn{3}{c}{\tiny a huge car is turning right side of the road near the building area}\\
   % \vspace{0.02cm}
    \adjustbox{trim={.1\width} {.22\height} {0.1\width} {.24\height},clip}
    {\includegraphics[width=0.4\linewidth]{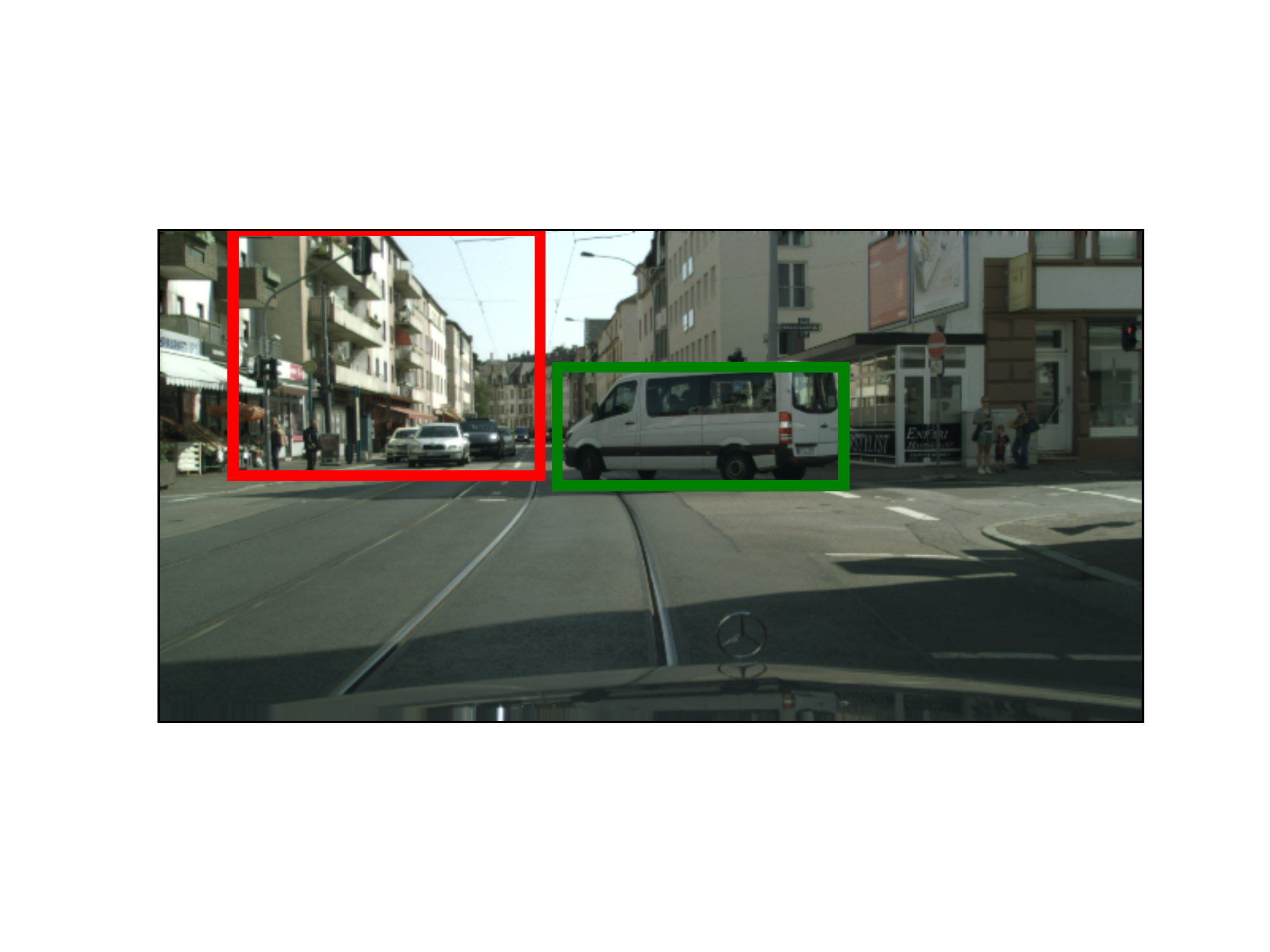}}
\noindent 
\adjustbox{trim={.1\width} {.22\height} {0.1\width} {.24\height},clip}
    {\includegraphics[width=0.4\linewidth]{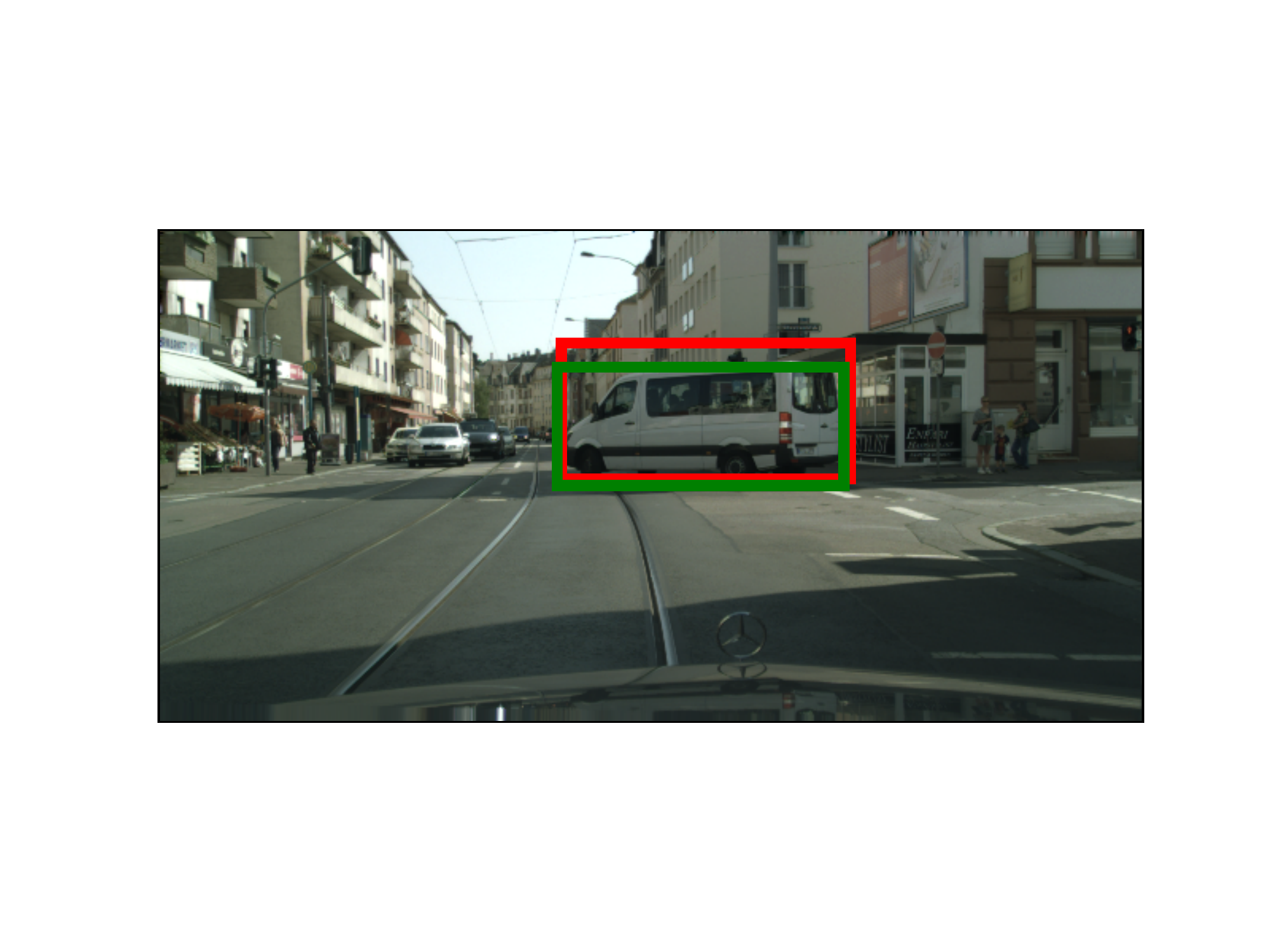}}
    \adjustbox{trim={.1\width} {.18\height} {0.1\width} {.20\height},clip}
    {\includegraphics[width=0.4\linewidth]{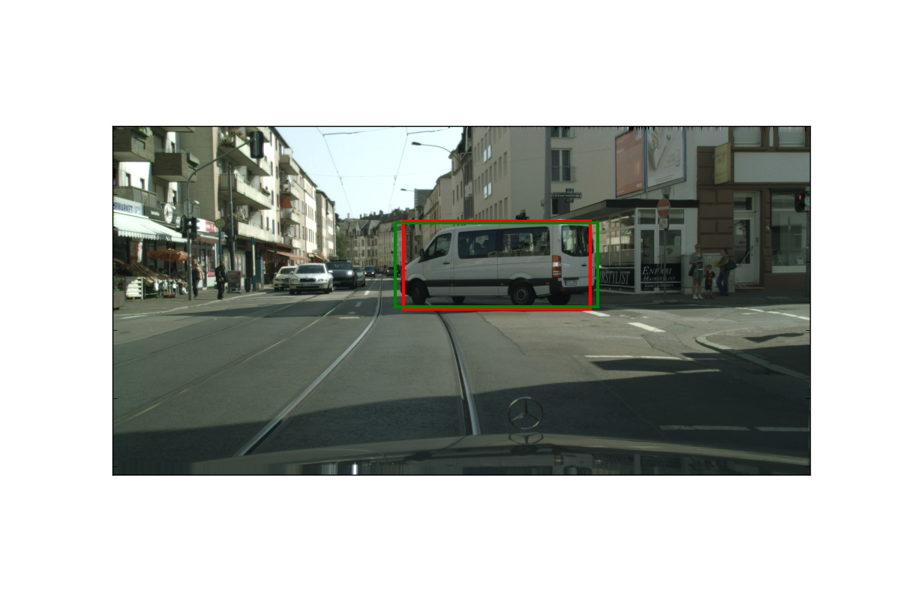}} \vspace{-1mm} \\
    \multicolumn{3}{c}{\tiny a woman in white dress in front is walking towards us on right side of road along with her family.}\\
        \adjustbox{trim={.1\width} {.2\height} {0.1\width} {.22\height},clip}
    {\includegraphics[width=0.4\linewidth]{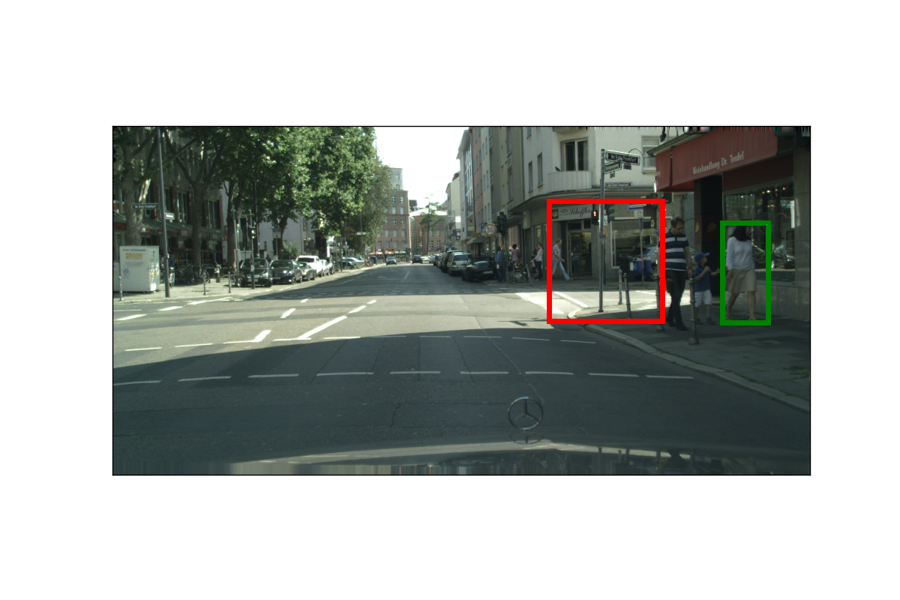}}
\noindent 
\adjustbox{trim={.1\width} {.2\height} {0.1\width} {.22\height},clip}
    {\includegraphics[width=0.4\linewidth]{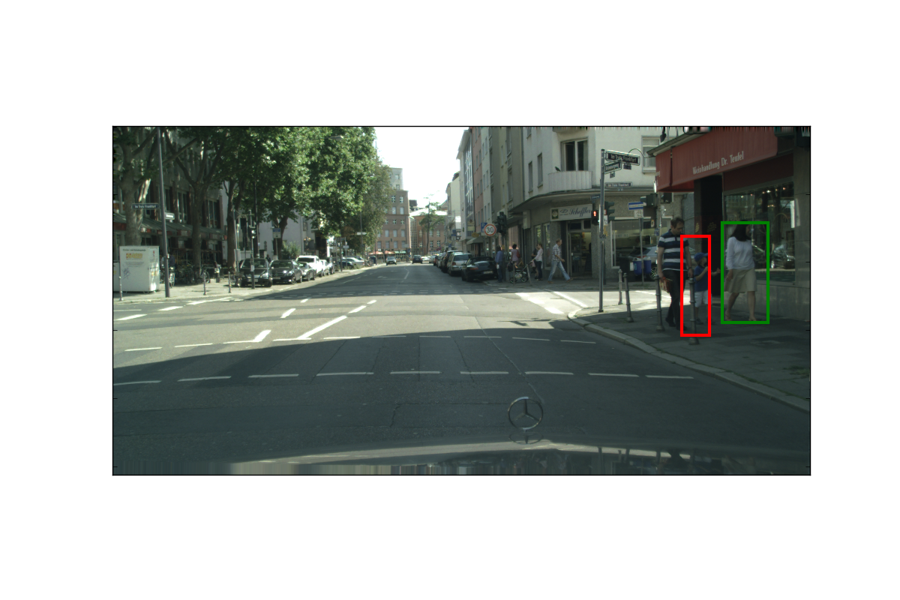}}
    \adjustbox{trim={.1\width} {.2\height} {0.1\width} {.22\height},clip}
    {\includegraphics[width=0.4\linewidth]{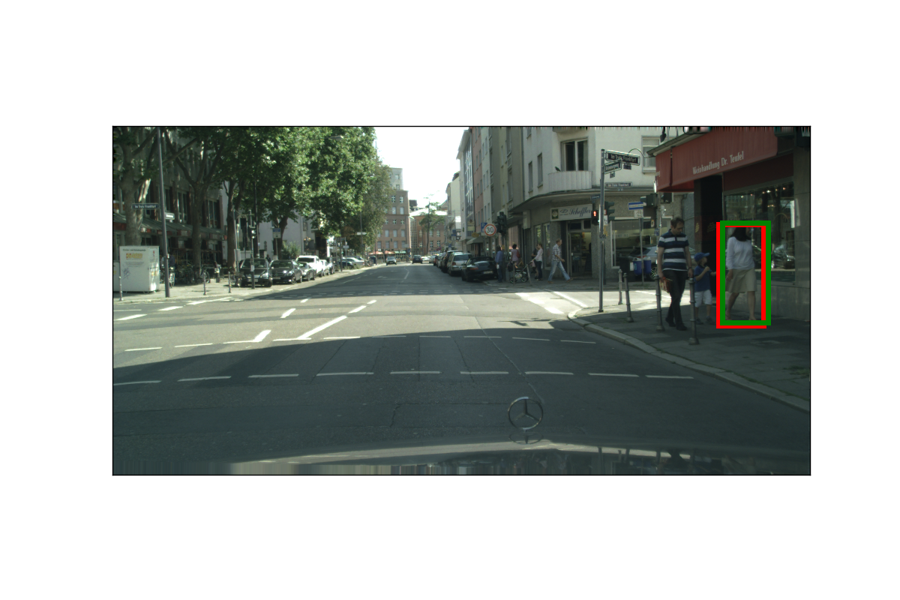}} \vspace{-1mm}\\ 
    \multicolumn{3}{c}{\tiny a car in front at a very far distance is moving on left side of the road}\\
            \adjustbox{trim={.1\width} {.2\height} {0.1\width} {.22\height},clip}
    {\includegraphics[width=0.4\linewidth]{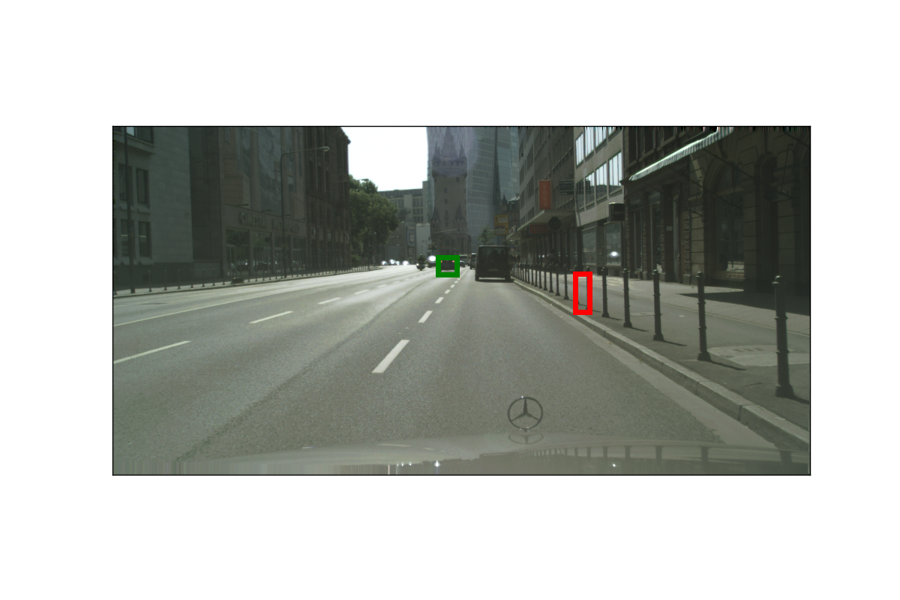}}
\noindent 
\adjustbox{trim={.1\width} {.2\height} {0.1\width} {.22\height},clip}
    {\includegraphics[width=0.4\linewidth]{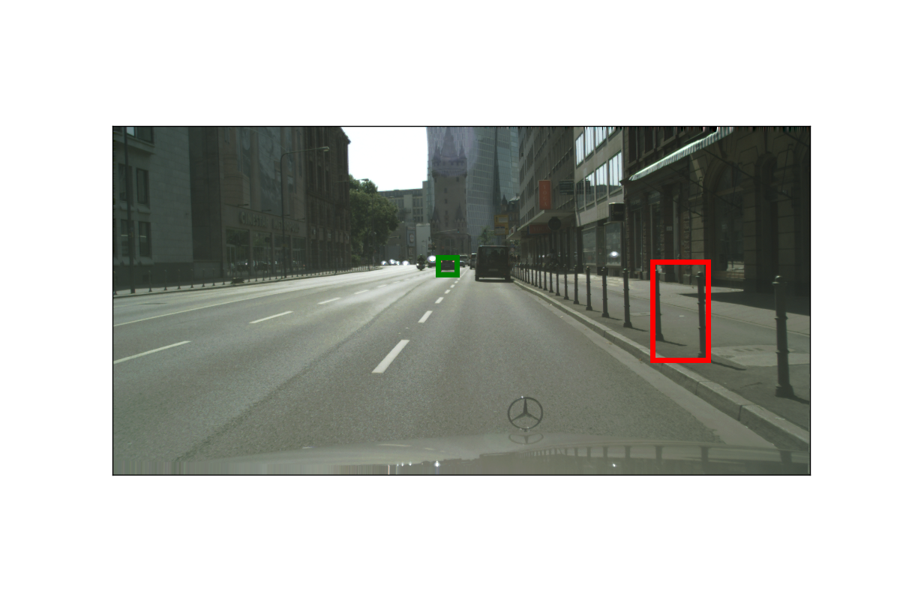}}
    \adjustbox{trim={.1\width} {.2\height} {0.1\width} {.22\height},clip}
    {\includegraphics[width=0.4\linewidth]{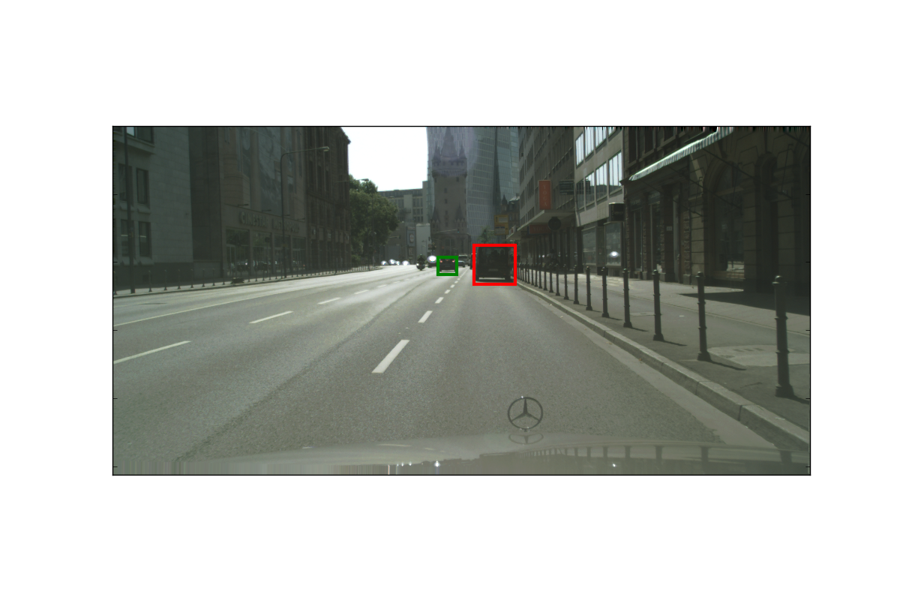}}\\
%\noindent 
%\includegraphics[width=0.5\linewidth]{figures/Recall-IoU-100proposals.png}
%\noindent 
%\includegraphics[width=0.5\linewidth]{figures/Recall-Candidates-IoU_0_9.png}
\end{tabular}
\caption{Some qualitative results from: NLOR (left column), Ours(I,D,O) (middle) and Ours(I,D,O,G) (right column). These results are obtained on the Cityscapes.  {\color{green} Green}: ground truth box and {\color{red}Red}: predicted box.}
\label{fig:qual-results-cityscape}
\vspace{-1mm}
\end{figure}
%---------------------------------------------------

%-----------------------------------------------------
\begin{figure*} 
  \centering
   $ \begin{array}{cccccc}  \hspace{-4mm}
 \begin{turn}{90}{\scriptsize{Descriptions}}\end{turn} & \hspace{-3mm}
\begin{tcolorbox}[width=0.24\linewidth, left=1pt,right=1pt,top=0pt,bottom=0pt]
a woman in front is crossing the road from left to right side with a travel bag
\end{tcolorbox} 
& \hspace{-2mm}
\begin{tcolorbox}[width=0.24\linewidth, left=1pt,right=1pt,top=0pt,bottom=0pt]
a red car in front is parked on right side of road along with other cars             
\end{tcolorbox} 
& \hspace{-2mm}
\begin{tcolorbox}[width=0.24\linewidth, left=1pt,right=1pt,top=0pt,bottom=0pt]
a woman in front is crossing the road from right to left with others
\end{tcolorbox} 
& \hspace{-2mm}
\begin{tcolorbox}[width=0.24\linewidth, left=1pt,right=1pt,top=0pt,bottom=0pt]
a man in front is walking on the left side of another road
\end{tcolorbox} \vspace{-4.5mm} \\ \hspace{-4mm}
 \begin{turn}{90}{\scriptsize{Video and Gaze}}\end{turn}
\hspace{0.6mm} & \hspace{-3mm}
\includegraphics[width=0.24\linewidth]{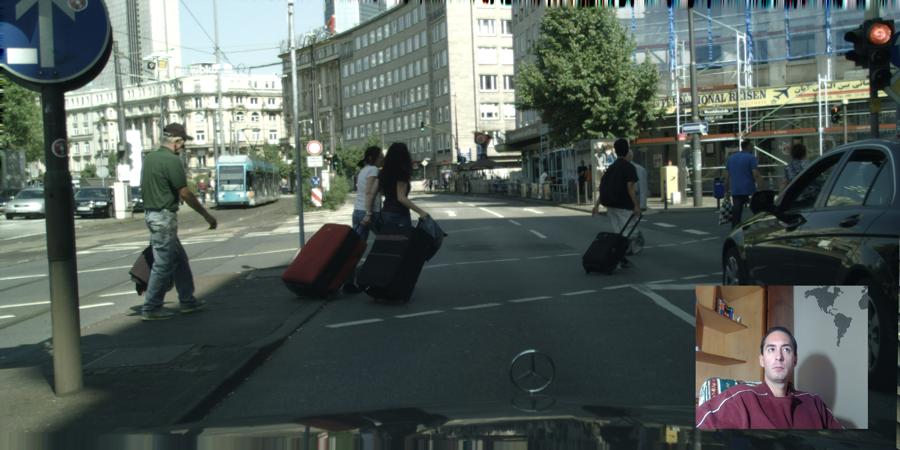} & 
    \hspace{-2mm}
\includegraphics[width=0.24\linewidth]{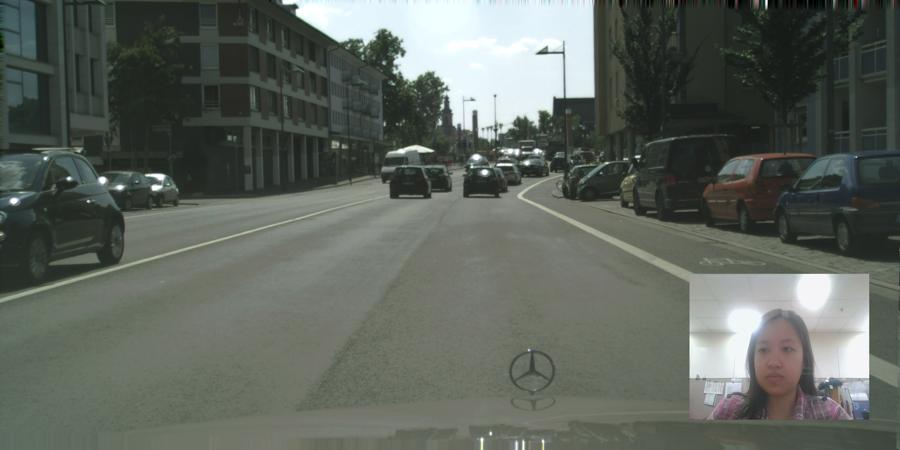} &  
    \hspace{-2mm}
 \includegraphics[width=0.24\linewidth]{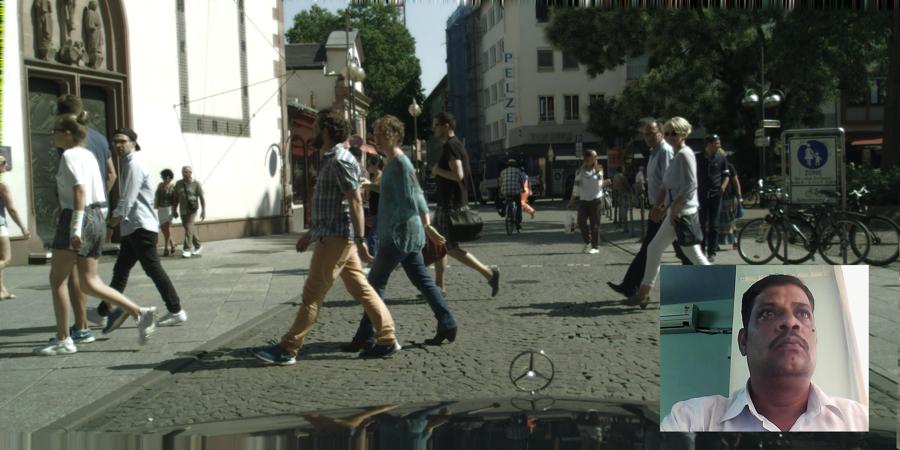} & 
    \hspace{-2mm}
\includegraphics[width=0.24\linewidth]{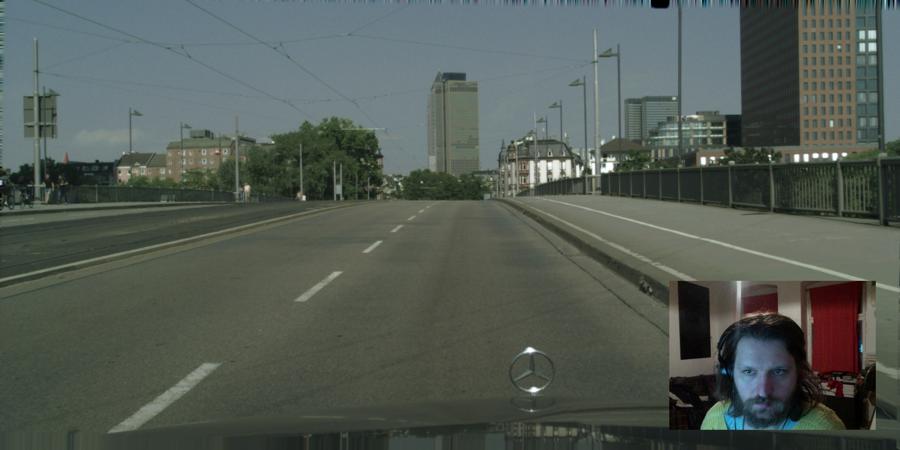} \vspace{1.5mm} \\ \hspace{-4mm}

 \begin{turn}{90}{\scriptsize{Intermediate Results}}\end{turn}
 \hspace{0.6mm}
&  \hspace{-3mm}
     \includegraphics[width=0.24\linewidth]{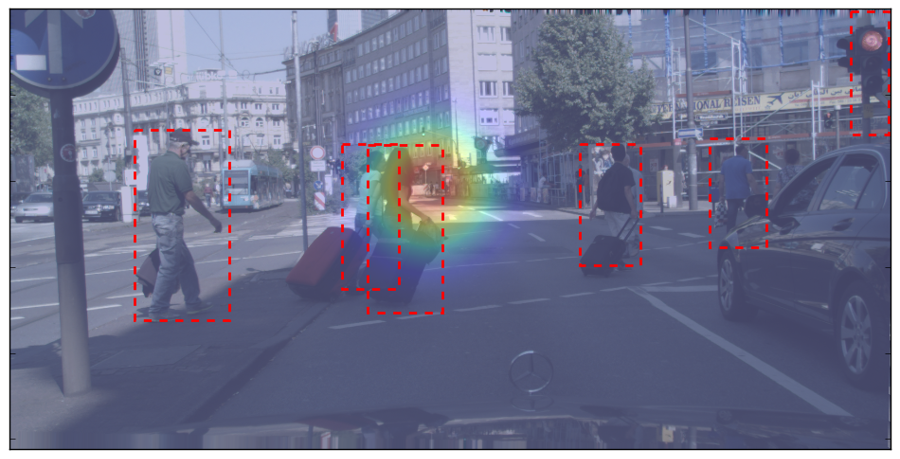}
 
& \hspace{-2mm}
     \includegraphics[width=0.24\linewidth]{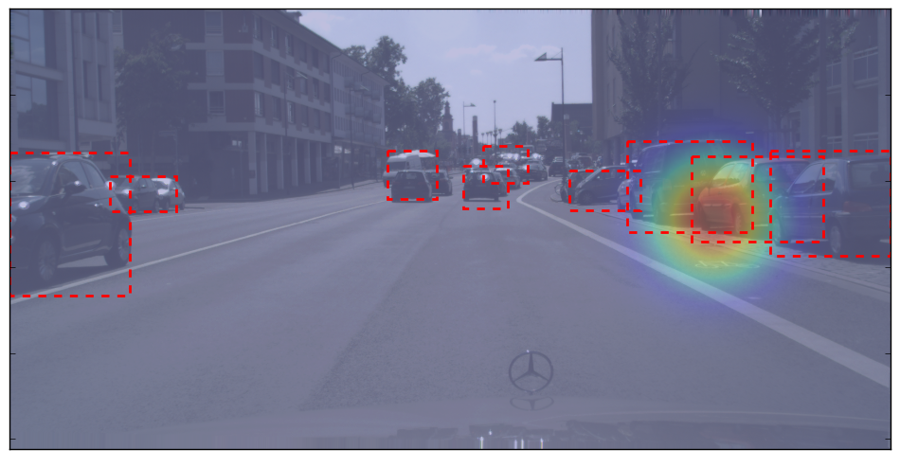}

& \hspace{-2mm}
     \includegraphics[width=0.24\linewidth]{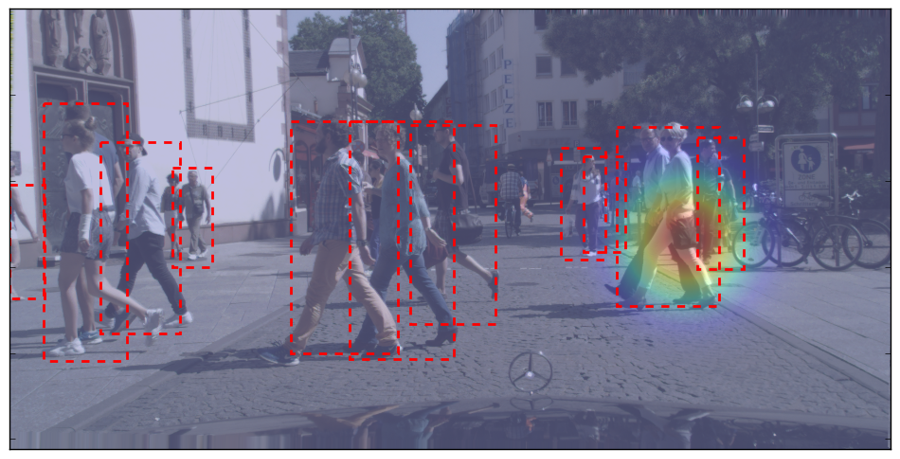}
    
& \hspace{-2mm}
    \includegraphics[width=0.24\linewidth]{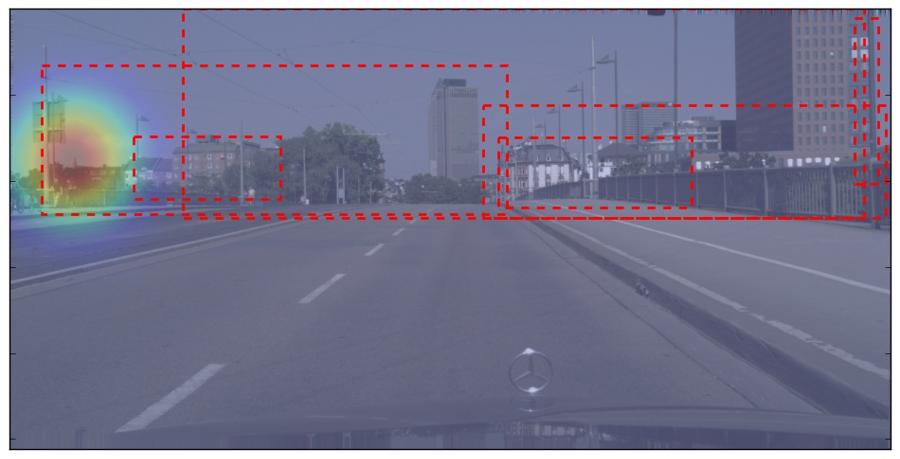} \vspace{1.2mm}
\\ \hspace{-3mm}
 \begin{turn}{90}{\scriptsize{Final Results}}\end{turn}
 \hspace{0.6mm}  
& \hspace{-3mm} \includegraphics[width=0.24\linewidth]{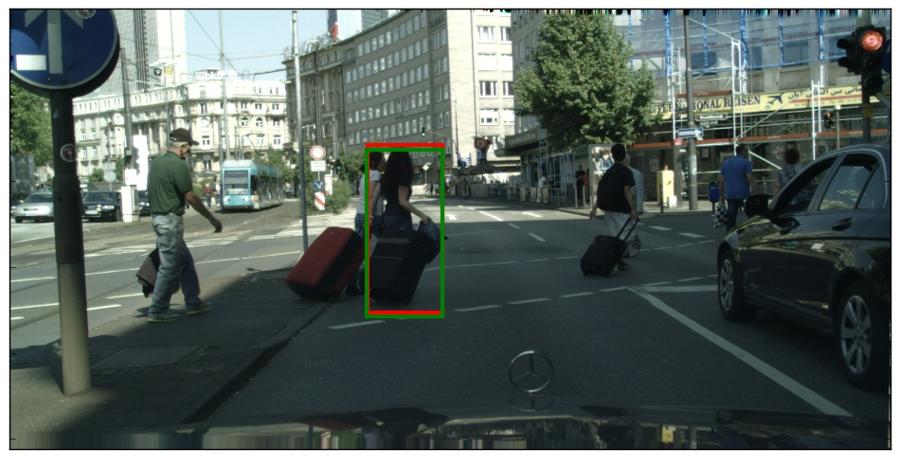}

& \hspace{-2mm} \includegraphics[width=0.24\linewidth]{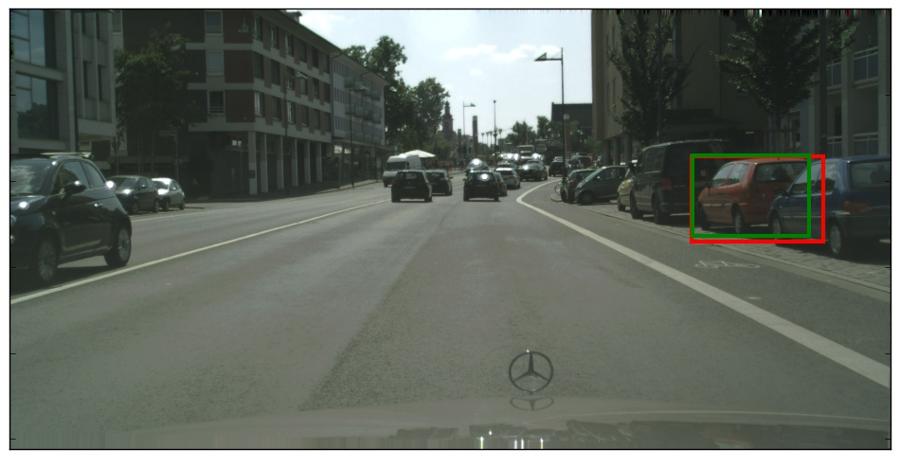}

& \hspace{-2mm} \includegraphics[width=0.24\linewidth]{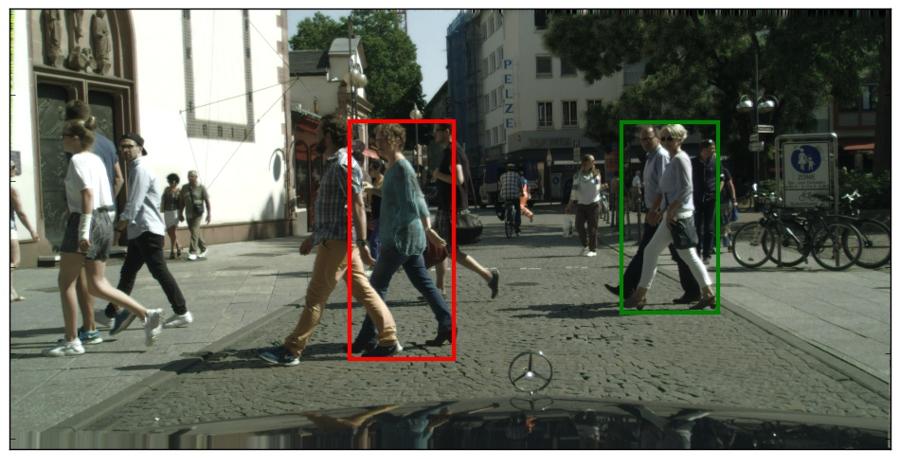}

& \hspace{-2mm} \includegraphics[width=0.24\linewidth]{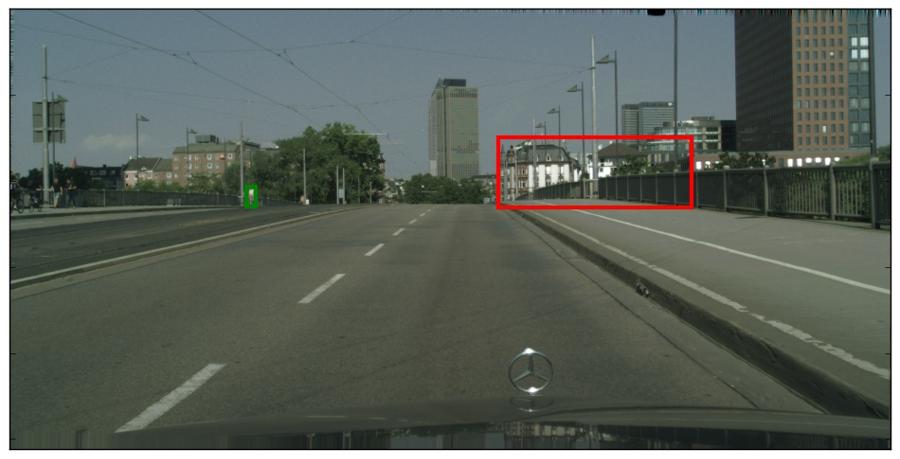}
\\
\end{array}$
\caption{Overall results on Cityscapes. Input descriptions on the first row and input videoframe and gaze in the second row. Middle row represents intermediate results where Gaze estimation is embedded along with object proposals while bottom row represents the final OR results. {\color{green} Green}: ground truth box and {\color{red}Red}: proposals and the predicted boxes.}
\label{fig:qual-gaze-results-cityscape}
\end{figure*}
%-----------------------------------------------

\bigskip
\noindent
\textbf{Object Proposal}
We use object proposal methods, e.g. Edgebox~\cite{ZitnickECCV14edgeBoxes}, FRCNN~\cite{renNIPS15fasterrcnn} and LOP~\cite{vasudevan2017chi} and compare them in Tab.~\ref{tab:eop-expt}. Since LOP performs consistently better for OR, we use the same for all later experiments.

\bigskip
\noindent
\textbf{Images vs. Stereo Videos}.
The performance of our model using different modalities and their combinations is reported in Tab.~\ref{tab:eop-expt}. 
We compare with better performing discriminative approaches (compared to CCA~\cite{plummer2015flickr30k} as in ~\cite{kiros2014unifying,liu2015multi}) like MNLM~\cite{kiros2014unifying}, VSEM~\cite{liu2015multi}, MCB~\cite{multimodal:pooling} as in Tab.~\ref{tab:eop-expt}. Surprisingly, the above approaches perform worse than NLOR~\cite{hu2016natural}, which uses a simple generative model. This could be because discriminative methods are more dataset-dependent and harder to generalize to new datasets/tasks. For instance, they expect `carefully-engineered' \textit{negatives} selection and sometimes more structured language expressions~\cite{plummer2015flickr30k}. 
We choose \cite{hu2016natural} as baseline due to its simplicity and it only requiring \textit{positives}. 
The table shows that our model can effectively utilize additional modalities such as the depth and motion provided in stereo videos.  
For instance, the $Acc@1$ is improved by $5.105$\% by using depth and motion; that is the improvement of Ours (I,D,O) over NLOR under LOP for 30 object proposals (I:RGB, D:Depth map, O:Optical Flow, G:Gaze). 
We observe a similar improvement for FRCNN as object proposals. Since LOP is consistent over all methods, we choose LOP as proposal technique for Tab.~\ref{tab:cityscape-track} and Tab.~\ref{tab:gaze-comparison}. We observe that both depth and motion can boost OR performance, on top of RGB appearances alone. For example, depth improves the performance by $2.743$\% and motion improves it by $3.855$\%. The combination of all four (appearance, depth, motion and gaze) yields the best results improving by $8.367$\%, indicating its usefulness for the OR task. Please see Fig.~\ref{fig:qual-results-cityscape} for some visual results and how the three additional modalities (D,O,G) improve the detections over the original work~\cite{hu2016natural}.

Also according to the table, motion features are more useful than depth in our case. This can be ascribed to the fact that the convolutional network to extract motion was trained with a larger dataset than for the depth network. A better representation of depth than HHA images is a good topic for further research. 
In Tab.~\ref{tab:cityscape-track}, we included flow information from the past by experimenting with different track lengths. We see that longer tracks do not bring significant improvement to OR accuracy. This may be because a) the referring expressions are annotated for the last frame (referring expressions are likely to be valid only for a short time as both camera and objects may be moving.~\eg \textit{the women \textbf{entering} the door}) and b) the length of Cityscapes videos is just 2 sec., too short to contain significant motion changes. 

\bigskip
\noindent
\textbf{Gaze vs. w/o Gaze}.
Gaze features are given to our model as additional local features along with depth and flow features. Comparing our model with and without Gaze, Gaze improves its performance significantly for all its variants (Tab.~\ref{tab:gaze-comparison}). For instance, Gaze features under Max pooling improve the performance by $3.773$\% for the image-only case (Ours (I)); by $1.748$\% when image and motion are used (Ours (I,O)); and by $3.262$\% when image, motion and depth are used (Ours (I,D,O)). Human gaze consistently improves OR performance, because humans do gaze at objects when issuing referring expressions. 

Given sampling rate differences between a Cityscapes video and its gaze video, we experimented with gaze feature extraction in 2 cases: a) timestamp matching between the videos, b) \# gaze frames $>$ \# frames in the Cityscapes video, where we tried average and max pooling of object features to ensure one-to-one correspondence between frames. Tab.~\ref{tab:gaze-comparison} shows that max pooling of object features performs as good or better than other cases such as averaging pooling because errors due to quickly changing gaze (outliers) can be avoided by max-pooling the object features.

\bigskip
\noindent
\textbf{Qualitative Results}.
We provide qualitative results in Fig.~\ref{fig:qual-gaze-results-cityscape}. The top row represents the inputs, incl. a Cityscapes video, a gaze recording video and the referring expression. Having overlaid the gaze estimate over object proposals (middle row), we can also observe the proximity of the gaze estimate to the referred objects. We show the predicted and groundtruth boxes in the bottom row. We add some failure cases in Fig.~\ref{fig:qual-gaze-results-cityscape}. From the above experiments, we infer that using multi-modal inputs - depth, motion, and gaze, along with RGB image features - improves OR performance.

\section{Conclusions} 
\label{sec:conclusion}

In this work, we have proposed a solution for object referring (OR) in videos using language and speaker's gaze. The main contributions are: 1) a new video OR dataset with $30,000$ objects annotated across $5,000$ different video sequences; 2) a novel approach Temporal-Spatial Context Recurrent ConvNet for OR in videos, which integrates appearance, motion, depth, human gaze and spatio-temporal context that can be trained in an end-to-end fashion; and 3) gaze recordings for all annotated objects and demonstration of their effectiveness for OR. 
Experiments show that our model can effectively utilize motion cues, gaze cues and spatio-temporal context provided by stereo videos, outperforming image-based OR methods consistently. Training and evaluating our method, especially the contribution of the multiple modalities, in a real human-to-robot communication system are future works.

\noindent
\textbf{Acknowledgement}: The work is supported by Toyota via project TRACE-Zurich. We also thank Vaishakh Patil and Prashanth Chandran for helpful comments. 

{\small
\bibliographystyle{ieee}
\bibliography{egbib}
}

\end{document}